\documentclass[runningheads]{llncs}
\usepackage[T1]{fontenc}
\usepackage{graphicx}
\usepackage{soul} % for highlighting
\usepackage{multirow}
\usepackage{bbm}
\usepackage{float}
\usepackage{makecell}
\usepackage{booktabs}
\usepackage{amsmath} 
\usepackage{hyperref}
\usepackage{xcolor}
\DeclareMathOperator*{\argmin}{arg\,min}
\usepackage{caption}
\usepackage[normalem]{ulem}
\usepackage{subcaption}
\usepackage[capitalize]{cleveref}

\definecolor{LightGreen}{RGB}{230,250,245} % Example RGB values for a special color
\definecolor{DarkGreen}{RGB}{0,120,87}
\definecolor{LightOrange}{RGB}{255,240,235} % Example RGB values for a special color
\definecolor{DarkOrange}{RGB}{212,82,41}

\newcommand{\highlightgreen}[1]{\textcolor{DarkGreen}{\setlength{\fboxsep}{0pt}\colorbox{LightGreen}{#1}}}
\newcommand{\highlightorange}[1]{\textcolor{DarkOrange}{\setlength{\fboxsep}{0pt}\colorbox{LightOrange}{\sout{#1}}}}
\newcommand{\highlightorangew}[1]{\textcolor{DarkOrange}{\setlength{\fboxsep}{0pt}\colorbox{LightOrange}{#1}}}

\newcolumntype{P}[1]{>{\centering\arraybackslash}p{#1}}

\newcolumntype {+}{ >{\global\let\currentrowstyle\relax}}
\newcolumntype {^}{ >{\currentrowstyle }}

\usepackage{xspace}
\newcommand{\cgg}{CGG\xspace}
\newcommand{\cgv}{CGV\xspace}
\newenvironment{courier}{%
    \fontsize{7}{7}\fontfamily{pcr}\selectfont 
}{%
    \par 
}
\usepackage{pifont}% http://ctan.org/pkg/pifont
\newcommand{\cmark}{\ding{51}\xspace}%
\newcommand{\xmark}{\ding{55}\xspace}
\usepackage{multirow}

\title{Guiding LLMs to Generate High-Fidelity and High-Quality Counterfactual Explanations for Text Classification}
\titlerunning{LLMs for Counterfactuals on Texts}

\author{Van Bach Nguyen\inst{1}\orcidID{0000-0002-4576-9302} \and
Christin Seifert\inst{1}\orcidID{0000-0002-6776-3868}  \and
Jörg Schlötterer\inst{1,2,3}\orcidID{0000-0002-3678-0390}}

\authorrunning{Van Bach Nguyen et al.}
\institute{University of Marburg, Germany
\and
University of Mannheim, Germany\\
\email{
\{vanbach.nguyen, christin.seifert, joerg.schloetterer\}@uni-marburg.de
}
}
\begin{document}
\maketitle
\begin{abstract}
The need for interpretability in deep learning has driven interest in counterfactual explanations, which identify minimal changes to an instance that change a model’s prediction. Current counterfactual (CF) generation methods require task-specific fine-tuning and produce low-quality text. Large Language Models (LLMs), though effective for high-quality text generation, struggle with label-flipping counterfactuals (i.e., counterfactuals that change the prediction) without fine-tuning.
We introduce two simple classifier-guided approaches to support counterfactual generation by LLMs, eliminating the need for fine-tuning while preserving the strengths of LLMs.  Despite their simplicity, our methods outperform state-of-the-art counterfactual generation methods and are effective across different LLMs, highlighting the benefits of guiding counterfactual generation by LLMs with classifier information.
We further show that data augmentation by our generated CFs can improve a classifier's robustness.
Our analysis reveals a critical issue in counterfactual generation by LLMs: LLMs rely on parametric knowledge rather than faithfully following the classifier.\footnote{Source code is available at \url{https://github.com/bach1292/llm-cls}}
%More important, our analysis shows that words deemed important by the classifier for prediction are not always the ones to modify for counterfactual explanations. Furthermore, we highlight the reliance of LLMs on parametric knowledge rather than contextual information or instructions for these tasks.\footnote{The source code is included with the paper submission and will be publicly accessible upon acceptance.}
\end{abstract}

\section{Introduction}
\begin{figure}[t]
\centering
  \includegraphics[trim=0 0 0 1.1cm, clip, width=0.7\columnwidth]{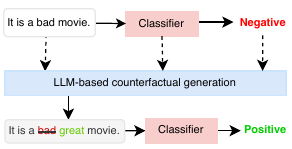}
  \caption{Counterfactual explanations to explain a text classifier. Counterfactuals (CFs) are minimal changes to the input, such that the classifier assigns a different label. 
  We propose and evaluate two approaches to generate CFs by LLMs that use information from the classifier (besides the original input and label) to guide the generation process post-hoc and ante-hoc.}
  \label{fig:intro}
\end{figure}

%Counterfactual explanations in context of eXplainable AI (XAI) are defined as texts with minimal changes to an input text that result in a changed prediction by a classifier. They can provide valuable insight into the classifier's decision-making process~\cite{miller-2019-explanation,wu-etal-2021-polyjuice}, thereby fostering interpretability \footnote{Counterfactuals in XAI context are different from real-world counterfactuals that describe hypothetical scenarios that contradict known facts~\cite{kulakova2013processing}}.
Counterfactual explanations identify minimal changes to the input that result in a changed prediction by a classifier. As such, they are a core tool for eXplainable AI (XAI), providing valuable insights into a classifier's decision-making process~\cite{miller-2019-explanation,wu-etal-2021-polyjuice}, thereby fostering interpretability. Counterfactual explanations in an XAI context differ from real-world counterfactuals: the latter describe hypothetical scenarios that contradict known facts~\cite{kulakova2013processing}, while the former aim to uncover a classifier's decision-making process.
For example, given a negatively classified movie review, identifying minimal changes (adding, replacing or deleting words) that would turn the classifier's prediction into positive, while maintaining meaningful content (cf.~\cref{fig:intro}) can provide insight into the classifier's reasoning and decision boundary. 
This insight can expose undesired model behavior, such as biases against minorities~\cite{mehrabi2021survey}, reliance on spurious correlations~\cite{geirhos2020shortcut}, or vulnerability to adversarial manipulations~\cite{goodfellow2014explaining}. 

\noindent Counterfactual generation methods that access the classifier produce counterfactuals (CFs) with a high label-flip rate (i.e., they succeed in actually changing the prediction) and minimal changes~\cite{ross-etal-2021-explaining,treviso-etal-2023-crest,robeer-etal-2021-generating-realistic}. However, they require task-specific fine-tuning and may generate low-quality text~\cite{nguyen-etal-2024-ceval-benchmark}.
In contrast, Large Language Models (LLMs) are used for end-to-end generation without expensive fine-tuning and produce high-quality text~\cite{radford2019language,kojima2022large}. However, they struggle to flip the labels with minimal changes~\cite{nguyen-etal-2024-ceval-benchmark,bach2024llms}, i.e., do not generate high-fidelity\footnote{The terms \emph{fidelity} and \emph{faithful} are sometimes used synonymously to describe how well an explanation aligns with the true reasoning of a classifier. In this paper, we use \emph{fidelity} to refer to ``true counterfactuals'', i.e., counterfactuals that actually change a classifier's prediction and \emph{faithful} to describe whether a counterfactual truly explains the classifier's actual reasoning.} CFs, because they do not have information about the classifier and its decision boundary.
In this paper, we aim to bridge the gap between fully supervised, costly CF generation methods and unsupervised, end-to-end text generation using LLMs.
We present two simple yet effective approaches for incorporating classifier information to guide LLMs in CF generation (cf.~\cref{fig:intro}), enabling high-quality text generation without the need for fine-tuning and offering broad applicability. 
In an ante-hoc approach, we first identify important words using an XAI method (saliency maps~\cite{simonyan2013deep} and SHAP~\cite{lundberg2017shap}). We then use the most important words and examples of factual-counterfactual pairs to prompt the LLM to generate counterfactual texts. Our second approach prompts the LLM to generate a set of candidate counterfactuals, evaluates their fidelity to the classifier, and selects the best candidate post-hoc based on minimality of changes and fidelity.

Specifically, the contributions of this paper are:
\begin{itemize}
    \item We show that LLMs can generate high-fidelity explanations with simple guidance signals, and can outperform fully-supervised state-of-the-art methods.
    \item We find that this performance depends on parametric knowledge of the LLM. LLMs generate high-fidelity counterfactuals for accurate classifiers, i.e., the counterfactuals reflect counterfactuals in the real-world, rather than faithfully explaining the classifier's true reasoning.
    \item Using the generated counterfactuals for data augmentation, we show that LLM-guided counterfactuals are an effective method to improve the robustness of text classifiers.
\end{itemize}

\section{Related work}
Counterfactual text generation methods fall into three categories~\cite{nguyen-etal-2024-ceval-benchmark}: masking and filling approaches, sampling based on conditional distributions, and LLM-based generation. 

\paragraph{Masking and Filling (MF):} These methods involve (1) masking key words and (2) filling the masked words using a pretrained language model. MICE~\cite{ross-etal-2021-explaining} and AutoCAD~\cite{wen-etal-2022-autocad} use classifier gradients to identify important words, while DoCoGen~\cite{calderon-etal-2022-docogen} masks n-grams with scores exceeding a threshold. CREST~\cite{treviso-etal-2023-crest} trains SPECTRA~\cite{guerreiro-martins-2021-spectra} to detect masking candidates. In step (2), these methods fine-tune T5~\cite{raffel2020exploring} to fill the blanks. Polyjuice~\cite{wu-etal-2021-polyjuice} allows user-defined masking and fine-tunes a RoBERTa-based model using control codes to fill the blanks.

\paragraph{Conditional Distribution (CD):}CF-GAN~\cite{robeer-etal-2021-generating-realistic} and CLOSS~\cite{fern-pope-2021-text} model a conditional distribution to generate counterfactuals based on the target label. CF-GAN adapts StarGAN~\cite{choi2018stargan} to produce realistic counterfactuals. CLOSS generates text counterfactuals by optimizing token embeddings, ranking impactful changes with Shapley values, and using beam search to select minimal, grammatically plausible edits that flip a model’s prediction. 

\paragraph{LLM-Based Generation:} CORE~\cite{dixit-etal-2022-core} trains a counterfactual retriever using contrastive learning, where positive pairs consist of human-authored counterfactuals and negative pairs are paraphrases. The retriever selects relevant counterfactual excerpts from an unlabeled corpus, which are then incorporated into prompts to guide GPT-3~\cite{brown2020language} in generating diverse counterfactual edits. CORE relies on counterfactual data and does not use classifier information. In contrast, DISCO~\cite{chen-etal-2023-disco} generates a large set of perturbations by masking spans identified by a neural syntactic parser and filling them using GPT-3. A filtering model then selects the most promising counterfactuals. However, DISCO is computationally expensive and therefore primarily suited for short-text tasks such as natural language inference (NLI). Additionally, DISCO focuses on generating CFs for data augmentation rather than explanations. FLARE~\cite{bhattacharjee2024towards} generates CFs by prompting LLMs in three steps: extracting latent features, identifying relevant words linked to those features, and minimally modifying these words to produce CFs.

\noindent In summary, MF and CD methods require task-specific fine-tuning, while LLM-based generation lacks classifier integration and produces poor fidelity counterfactuals~\cite{nguyen-etal-2024-ceval-benchmark}. 
Our work not only addresses the limitations of both, fine-tuned and LLM-based methods, but also combines their strengths.

\section{Guided LLM-based Counterfactuals}

\begin{figure}[t]
    \centering
    % First subfigure
    \begin{subfigure}{\linewidth}
        \centering
        \includegraphics[width=\columnwidth]{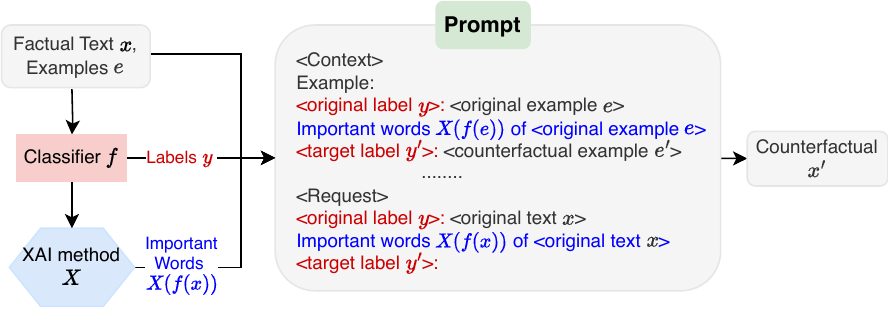}
        \caption{Classifier-Guided Generation (CGG)}
        \label{fig:method1}
    \end{subfigure}
\par\bigskip
    % Second subfigure
    \begin{subfigure}{\linewidth}
        \centering
        \includegraphics[width=\columnwidth]{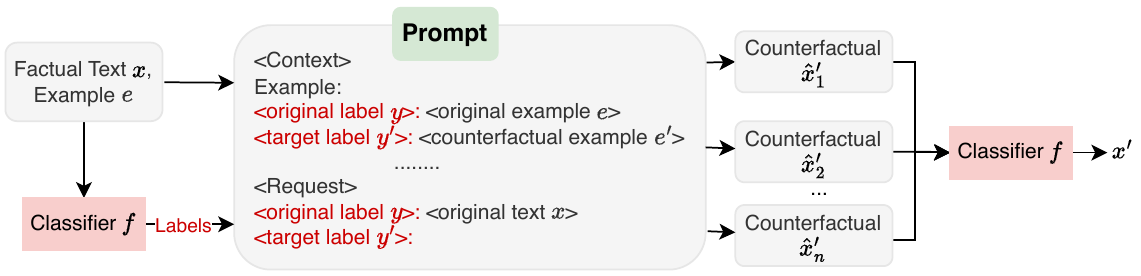}
        \caption{Classifier-Guided Validation (CGV)}
        \label{fig:method2}
    \end{subfigure}
    \caption{Overview of two classifier-guided approaches for generating counterfactual explanations. a) Classifier-Guided generation (CGG) uses an XAI feature importance method to identify relevant words for the prediction, which are used to extend the prompt for generating a counterfactual example with an LM. b) Classifier-Guided Validation (CGV) is a post-hoc method and selects the best counterfactual example from a set of (unguided) generated candidates.}
    \label{fig:methods}
\end{figure}

Given a classifier $f$ and an instance $x$, the original prediction is $f(x) = y$, and the counterfactual label is $y' \neq y$. Let $p$ be a prompt containing instructions, context, and an example using the classifier’s labels. $p(x, I)$ is the prompt applied to input $x$ with external information $I$. Given input prompt $p$, an LLM $M$ should generate a counterfactual $M(p) = x'$ such that $f(x') = y' \neq y$.

We propose two approaches that leverage information from the classifier $f$:  \textbf{C}lassifier-\textbf{G}uided \textbf{G}eneration (\textbf{\cgg{}}) and \textbf{C}lassifier-\textbf{G}uided \textbf{V}erification (\textbf{\cgv{}}). 
These two approaches guide LLMs by classifier information through two distinct mechanisms. In \cgg{}, the classifier’s information is used only during the generation process by providing important words and the classifier’s label for the example in the prompt, directly influencing the output. In \cgv{}, the classifier’s information is applied both during generation (by providing the classifier’s label for the example, similar to \cgg{}) and after generation (by using the classifier’s label to validate the output). The two approaches can be applied independently or in combination.

\subsection{Classifier-Guided Generation (CGG)}
In \textbf{\cgg{}}, the generation process is guided by information from the classifier. Specifically, we use an XAI method $X(f(x))$ to extract words from the input $x$ that are important to the classifier's decision $f(x)$. These important words are then incorporated into the predefined prompt $p$ to form $p(x, X(f(x)))$ (see \cref{fig:method1}). A complete example of the prompt is provided in~\cref{appx:prompts}. This prompt is input to the LLM $M$, which generates a counterfactual instance $x'$, such that $M(p(x, X(f(x)))) = x'$.

\subsection{Classifier-Guided Validation (CGV)}
\textbf{\cgv{}} uses the classifier after the generation process. Specifically, we use the LLM $M$ with prompt $p$ to generate $n$ candidate samples $\{\hat{x}_i\}_{i=1}^{n}$. The classifier $f$ is then used to validate the labels, assigning a label $\hat{y}_i$ to each sample, and we select the sample with the flipped target label $y'$ that also minimizes the distance $d(\hat{x}_i, x)$ to the original input $x$ (see \cref{fig:method2}).  Formally, we have
\[x' = \argmin_{\hat{x}_i: f(\hat{x}_i) = y'} d(\hat{x}_i, x).\]
Thus, $ x' $ is the sample with the desired label and minimal distance to the original input. If no sample flips the label, we select the instance with minimal distance to the input.
%In contrast, the \textbf{\cgv{}} approach uses the classifier after the generation process. Specifically, we use the LLM $M$ with prompt $p$ to generate $n$ samples $\{x'_i\}_{i=1}^{n}$. The classifier $f$ is then used to verify the labels, assigning a label $y'_i$ to each sample, and selecting the sample with the target flipped label $y'_{\text{target}}$ that also minimizes the distance $d(x', x)$ to the original input $x$ (see \cref{fig:method2}). Formally, we have
%\[x_{\text{selected}} = \arg \min_{x'_i: f(x'_i) = y'_{\text{target}}} d(x'_i, x).\]
%Thus, $ x_{\text{selected}} $ is the sample with the desired label and the minimal distance from the original input.

% We propose two approaches to utilize information from the classifier: Classifier-Guided Generation (\cgg{}) and Classifier-Guided Verification (\cgv{}). In this work, each approach is applied independently as well as in combination.

% In the \cgg{} approach, we use an explainable AI (XAI) method to extract important words from the classifier and integrate them into the prompts (see Figure~\ref{fig:method1}). The prompt is designed as a one-shot example using the classifier’s labels. The complete prompt is provided in Appendix A.

% In the \cgv{} approach, we generate \(n\) samples using a one-shot prompt with the classifier's label as an example. The classifier is then used to verify the label, selecting the sample with the flipped label that has the minimal distance from the original input (see Figure []).

\section{Experimental Setup}
We evaluate counterfactuals on the CEVAL benchmark~\cite{nguyen-etal-2024-ceval-benchmark},  which includes two datasets: IMDB~\cite{maas-etal-2011-learning}, a sentiment analysis dataset with binary labels (positive and negative), and SNLI~\cite{bowman-etal-2015-large}, a natural language inference dataset with three labels (entailment, contradiction, and neutral). The benchmark assesses performance by counterfactual and text quality metrics. 
A benchmark run required approximately two hours per dataset on an A100 80GB GPU.

\subsection{Models} Following the benchmark~\cite{nguyen-etal-2024-ceval-benchmark}, we use BERT-based classifiers $f$ that achieve 89\% accuracy on IMDB and 90\% on SNLI. We compare against MICE~\cite{ross-etal-2021-explaining}, the top method for counterfactual metrics according to the benchmark, CREST~\cite{treviso-etal-2023-crest}, a masking and filling method that uses SPECTRA~\cite{guerreiro-martins-2021-spectra} for masking and T5~\cite{raffel2020exploring} for filling in the blanks, and FLARE~\cite{bhattacharjee2024towards}, an LLM-based method that generates counterfactuals through three steps by prompting LLMs. The baseline is \textit{vanilla} 1-shot prompting without classifier guidance.

For the counterfactual generation LLM $M$, we compare Llama-3.1-8B-Instruct, Llama-2-7b-chat, and GPT-4o-mini\footnote{Version: GPT-4o-mini-2024-07-18.}. Following \cite{nguyen-etal-2024-ceval-benchmark}, we use a temperature of 1.0 for all LLMs in the counterfactual generation task.
For CGG, we adopt saliency maps~\cite{simonyan2013deep} as the XAI method $X$, selecting the top 25\% most important words with the highest gradients relative to the classifier’s prediction, consistent with the benchmark (additional results with SHAP~\cite{lundberg2017shap} are similar, cf. \cref{appx:shap}). For CGV, we use Levenshtein Distance as distance measure $d$ to select $x'$ and set the number of samples $n = 5$ for IMDB and $n = 10$ for SNLI, as larger $n$ values showed minimal differences in preliminary experiments. 
We use 1-shot prompts in our main experiments and analyze the impact of the number of shots with 5- and 10-shot prompts separately in section~\ref{sec:results:few_shot}).

\subsection{Evaluation Metrics}
We used the same set of metrics as in the CEVAL benchmark~\cite{nguyen-etal-2024-ceval-benchmark} to assess both counterfactual fidelity and text quality.

\subsubsection{Counterfactual metrics}
\begin{description}
    \item[\textbf{Flip Rate (FR)}] measures how effectively a method can change labels of instances with respect to a pretrained classifier $f$. FR is defined as the percentage of generated instances where the labels are flipped over the total number of instances $N$~\cite{bhattacharjee2024towards}:
\[FR =\frac{1}{N} \sum_{i=1}^{N} \mathbbm{1}[f(x_i) \neq f(x'_i)] \]
where $\mathbbm{1}$ is the indicator function.
\item[\textbf{Token Distance (Dis)}]
measures textual similarity by calculating the token-level Levenshtein distance \(d(x, x')\) between the original instance \(x\) and the counterfactual \(x'\). Levensthein distance is an ideal metric to assess minimality of edits as counterfactual generation involves specific edits (insertion, deletion, and substitution), rather than completely rewriting the text.
\[Dis= \frac{1}{N} \sum_{i=1}^{N} {d(x_i, x'_i})\]
\item[\textbf{Perplexity (PP)}]
evaluates whether the generated text is plausible, realistic, and follows a natural text distribution by calculating the perplexity of instance $x$ with the length of $n$ tokens using GPT-2:
\[PP(x) = \exp\left\{-\frac{1}{n} \sum_{i=1}^{n} \log p_{\theta}(z_i \mid z_{<i})\right\}\]
where \(\log p_{\theta}(z_i \mid z_{<i})\) is the log-likelihood of token \(z_i\) given the previous tokens \(z_{<i}\).
\end{description}

\subsubsection{Text quality metrics}
\begin{description}
    \item[\textbf{Grammar (GM)}] measures syntactical and grammatical accuracy.
    \item[\textbf{Fluency (Flu)}] assesses whether the text is readable and has a natural flow.
    \item[\textbf{Cohesiveness (Coh)}] assesses how logical and coherent the text structure is.
\end{description} 
Text quality metrics are assessed on a 5-point scale by prompting GPT-3.5 with a temperature of 0.2 in line with the benchmark~\cite{nguyen-etal-2024-ceval-benchmark}.

\section{Results}

We investigated the impact of guidance signals for LLMs on the quality of generated counterfactuals (Sec.~\ref{sec:results:performance}), analysed the impact of feature importance signals (Sec.~\ref{sec:results:featimp}), and the number of examples in few-shot prompting (Sec~\ref{sec:results:few_shot}).  
We used the generated examples to augment the training datasets (Sec.~\ref{sec:results:augment}) and investigated whether the generated counterfactuals are faithful to the classifier's true reasoning (Sec.~\ref{sec:results:parametric}).

\subsection{Incorporating Classifier Information Improves Performance on Counterfactual Metrics}
\label{sec:results:performance}
%\paragraph{Classifier Integration Significantly Improves Performance on Counterfactual Metrics}
Table \ref{tab:main_results} shows that our classifier-guided approaches improve key counterfactual metrics compared to the $vanilla$ baseline across both datasets and for all LLMs. 
\begin{table}[t]
\caption{Results for counterfactual metrics (FR: flip rate, Dis:  token-level Levensthein Distance, PP: perplexity) and text quality metrics (GM: grammar, Coh: Cohesiveness, Flu: Fluency) on IMDB and SNLI. \textit{Italic} denotes the best value within the same LLM, \textbf{bold} denotes the best value across all models, and the second-best value is \underline{underlined}. \textit{vanilla} indicates that no classifier information is used, and \textit{CGGV} denotes the combination of CGG and CGV.}
\label{tab:main_results}
\centering
 
\resizebox{\textwidth}{!}{
\begin{tabular}{clcccccccccccc}
\toprule
& & \multicolumn{6}{c}{\textbf{IMDB}} & \multicolumn{6}{c}{\textbf{SNLI}} \\
\cmidrule(lr){3-8} \cmidrule(lr){9-14}
& & \textbf{FR} $\uparrow$ & \textbf{Dis} $\downarrow$ & \textbf{PP} $\downarrow$  & \textbf{GM} $\uparrow$ & \textbf{Coh} $\uparrow$ & \textbf{Flu}  $\uparrow$
& \textbf{FR}$\uparrow$ & \textbf{Dis} $\downarrow$ & \textbf{PP} $\downarrow$ &  \textbf{GM} $\uparrow$ & \textbf{Coh} $\uparrow$ & \textbf{Flu}  $\uparrow$\\
\midrule
\multicolumn{2}{l}{FLARE}  & 0.89 & 39.4 & \underline{46} & 3.06 & 3.14 & 3.07 & 0.2 & \textbf{2.2} & 61 & 3.25 & 2.99 & 3.19 \\
\multicolumn{2}{l}{CREST}  & 0.71 & 70.5 & 45 & 2.18 & 1.27 & 2.33 & 0.39 & 3.5 & 61 & 2.71 & 2.74 & 2.70\\
\multicolumn{2}{l}{MICE} & \textbf{1.0} & 38.5 & 62 & 2.71 & 2.81 &  2.79 & \textbf{0.85} & 5.6  & 100 & 3.33 & 3.31 & 3.33 \\
\midrule
\multirow{4}{*}{\rotatebox{90}{\makecell{Llama-3.1}}}
& vanilla & 0.93 & 46.2 & \underline{\textit{46}} & \textit{\underline{3.10}} & \textit{\textbf{3.19} }&  \textit{\textbf{3.16}} & 0.45 & 5.3  & \textit{\underline{51}} &  \textit{3.63} & 3.59 & 3.59 \\
& \cgg{} & 0.94 & 45.4 & 56 & 2.85 & 2.93 & 2.95 & 0.40 & 4.4  & 58 & 3.62 & \textit{3.69 }& \textit{3.64} \\
& \cgv & \underline{0.99} & 34.6 & 48 & 3.00 & 3.07 &  3.05 & \textit{\underline{0.84}} & 3.7 & 55 & 3.57 & 3.55 & 3.60 \\
& CGGV & \textit{\textbf{1.0}} & \textit{\textbf{32.7}}& 55 & 2.80 & 2.89 & 2.91 & 0.73 & \textit{3.3}  & 59 & 3.56 & 3.53 & 3.57 \\
\midrule
\multirow{4}{*}{\rotatebox{90}{\makecell{Llama-2}}} & vanilla & 0.78  & 81.9 & 59 & \textit{\textbf{3.21}} & \textit{\underline{3.15}} & 2.70 & 0.41 & 5.2 & \textbf{\textit{50}}  & \textit{\textbf{3.92}} & 3.90 & 3.91\\
&\cgg& 0.86  & 90.6 & 60 & 3.05 & 3.06  & \textit{3.03} &  0.40 & 6.4 & 54 & \textit{\textbf{3.92}} & \textit{3.91} & \textbf{\textit{3.95}}  \\
&\cgv& 0.94  & 65.5 & \textbf{\textit{45}} & 2.84 & 2.90  & 2.89 &  \textit{0.70} & \textit{4.0} & 56 & 3.84 & 3.83 & 3.81  \\
& CGGV & \textit{0.97}  & \textit{64.0} & 68  & 2.91 & 2.96 & 2.94 &  0.65  & 5.0 & 56 & 3.89 & \textit{3.91} & 3.87  \\
\midrule
\multirow{4}{*}{\rotatebox{90}{\makecell{ GPT-4o-m}}}& vanilla & 0.94 & 47.5 & \textbf{\textit{45}} & \textit{\underline{3.10}} & \textit{\underline{3.15} }& \textit{\underline{3.10}} &  0.50 & 3.7 & \textit{58} & \textit{\underline{3.90}} & \underline{3.93} & \underline{\textit{3.92}}  \\
&\cgg & 0.97 & 43.9 & 54 & 2.93 & 3.01 & 3.03 &  0.44 & 3.8 & 62 & 3.71 & 3.79 & 3.76  \\
&\cgv& \underline{\textit{0.99}} & \underline{\textit{34.5}} & 47 & 3.06 & 3.12 & \textit{\underline{3.10}} &  \textit{0.75} & \underline{\textit{3.1}} & 61 & 3.87 & \textit{\textbf{3.95}} & \underline{\textit{3.92}}  \\
& CGGV & \underline{\textit{0.99}} & 36.6 & 54 & 2.87 & 2.99 & 2.97 &  0.63 & \underline{\textit{3.1}} & 64 & 3.67 & 3.72 & 3.72  \\
\bottomrule
\end{tabular}
}
\end{table}
For the combination of our two approaches (CGGV), we observe a consistent decrease of distance scores (Dis) and increase in flip rate (FR) in both datasets and across all LLMs.
Among the language models in the combination of our two approaches (CGGV), Llama-3.1 achieves the best results, outperforming the flip rate of CREST and FLARE and closely matching MICE, the state-of-the-art counterfactual generation method in terms of flip rate. 
Notably, with more shots, Llama-3.1 even surpasses MICE in FR on the SNLI dataset (see \cref{sec:results:few_shot}). 
These results are particularly encouraging, as CREST, FLARE and MICE are dedicated counterfactual generation methods that require task-specific fine-tuning. In contrast, despite their simplicity, our approaches are equally (or even more) effective in generating high-fidelity counterfactuals.
While text quality metrics (Grammar: GM, Fluency: Flu, Cohesiveness: Coh) show a slight decline -- likely due to distribution shifts from forced word modifications --  our approaches still outperform all other methods on these metrics except for FLARE on the IMDB dataset where performance is close. The high performance of CGV highlights that LLMs can generate high-quality counterfactuals with simple prompting, requiring only classifier-provided labels for guidance. 
\begin{table*}[ht!]
\caption{Examples  where LLMs succeed (\cmark) or fail (\xmark) to flip the label by changing the important words (highlighted in bold). }
\label{tab:examples}
\begin{tabular}{p{1.6cm}p{8.2cm}p{0.1cm}l}
\toprule
\textbf{Label} & \textbf{Text} & & \textbf{Model} \\
\midrule
\multicolumn{4}{c}{Natural Language Inference (SNLI)}\\\midrule
\highlightorangew{entailment} $\rightarrow$ \highlightgreen{neutral} & 
A young man \highlightorange{\textbf{standing}} \highlightgreen{stands} outside \highlightorange{a \textbf{laundromat}}. 
\newline
A man is standing. & & \xmark Llama-2 \\[1.8em]
\highlightorangew{entailment} $\rightarrow$ \highlightgreen{neutral} & A smiling man and a baby girl posing for photo. 
\newline
A girl \highlightorange{\textbf{poses} for a \textbf{selfie} with her mother} \highlightgreen{is smiling and} \highlightgreen{holding a toy}. & & \cmark Llama-2 \\[2.8em]
 \highlightorangew{entailment} $\rightarrow$ \highlightgreen{neutral} & A man in \highlightorange{\textbf{gray}} \highlightgreen{black} on a rocky cliff, overlooking the \highlightorange{\textbf{mountains}} \highlightgreen{ocean}.
 \newline
 A man can see mountains from where he stands. &  & \cmark Llama-3.1 \\
% \highlightorangew{entailment} $\rightarrow$ \highlightgreen{neutral}  & The girl \highlightorange{wearing a \textbf{brown} \textbf{jacket} whilst walking in \textbf{snow}} \highlightgreen{is walking outside}. The girl is walking outside. & Llama-2 (fail)
%  \\\\
% \highlightorangew{contradiction} $\rightarrow$ \highlightgreen{neutral}  & \highlightorange{\textbf{kids}}\highlightgreen{Kids} are looking at something \highlightorange{on a \textbf{table}}. Some kids are burning the table down. & Llama-2 (fail)\\\\
\midrule
\multicolumn{4}{c}{Sentiment Analysis (IMDB)}\\\midrule
\highlightorangew{Negative} $\rightarrow$\highlightgreen{Positive}  &  ...Did they even \highlightorange{\textbf{try}} \highlightgreen{consider} someone like Danny Devito?!? ....OK , now get this they cast Jessica Simpson did anyone take a \highlightorange{\textbf{look}} \highlightgreen{glance} at her husband? He \highlightorange{\textbf{matches}} \highlightgreen{fits} Luke Duke to a tee!!!!!!... &  & \xmark GPT-4o-m \\[3.8em]
 \highlightorangew{Negative} $\rightarrow$\highlightgreen{Positive} & being a NI \highlightorange{\textbf{supporter}} \highlightgreen{advocate}, it's \highlightorange{hard} \highlightgreen{easy} to \highlightorange{\textbf{objectively}} \highlightgreen{passionately} \highlightorange{\textbf{review}} \highlightgreen{enjoy} a movie \highlightorange{glorifying} \highlightgreen{celebrating} \textbf{ulster}  \highlightorange{\textbf{nationalists}} \highlightgreen{patriots}...&  & \cmark GPT-4o-m \\[2.8em]
 \highlightorangew{Negative} $\rightarrow$\highlightgreen{Positive}  & ... I \highlightorange{\textbf{hate}} \highlightgreen{have} to \highlightorange{down \textbf{talk}} \highlightgreen{give a shout-out to} another filmmaker so I'll just \highlightorange{\textbf{use constructive criticism}} \highlightgreen{share some praise}. 1. Find \highlightorange{\textbf{good}} \highlightgreen{talented} actors. Take the time. It \highlightorange{really \textbf{helps}} \highlightgreen{makes a huge difference}. ... &  & \xmark Llama-3.1 \\

\bottomrule
\end{tabular}
\end{table*}
\subsection{Not All LLMs Benefit Equally from Feature Importance Guidance Signals}
\label{sec:results:featimp}
\begin{figure}[t]
\centering
    \begin{subfigure}[h]{0.49\linewidth}
    \centering
        \includegraphics[width=\linewidth]{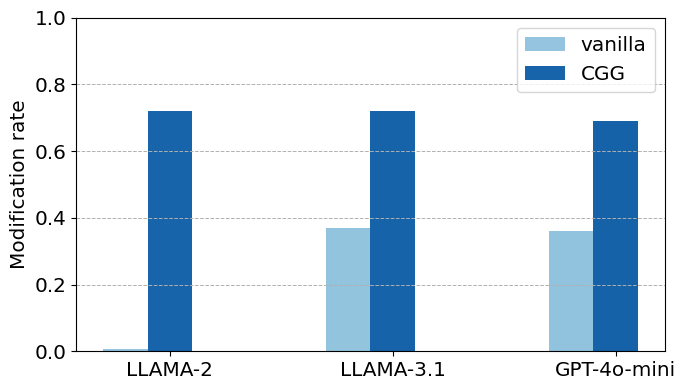}
        \caption{Impact of CGG on MR.}
        \label{fig:analysis_1}
    \end{subfigure}
\hfill
  \begin{subfigure}[h]{0.49\linewidth}
  \centering
        \includegraphics[width=\linewidth]{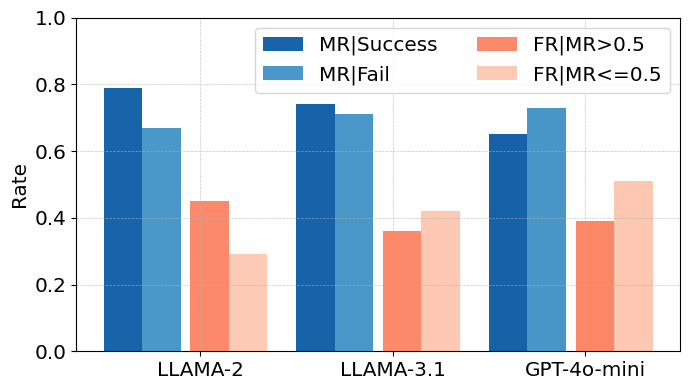}
        \caption{Dependencies between MR and FR}
        \label{fig:analysis_2}
    \end{subfigure}
  \caption{Impact of \cgg{} on modification rate (MR) and flip rate (FR) in the SNLI dataset. \cref{fig:analysis_1} illustrates the MR for each LLM, with and without CGG. \cref{fig:analysis_2} compares the MR for flipped (Success) and non-flipped (Fail) instances, alongside the FR for cases where MR $>$ 0.5 and MR $\leq$ 0.5.}
  \label{fig:modify_rate}
\end{figure}

We investigate whether LLMs effectively use the words identified as important by the XAI method in CGG by measuring the \textit{modification rate} (\textbf{MR}). MR is the percentage of important words that were changed, i.e., words present in the original text but missing from the generated text. As shown in \cref{fig:analysis_1}, CGG increases MR significantly from 0–40\% to 70–75\%. 
However, \cref{fig:analysis_2} shows that high MR does not guarantee flipping labels. While in LLama models, instances with flipped labels (true counterfactuals) are associated with higher MR than non-flipped instances, the opposite holds for GPT-4o-mini. Inspecting the flip rate for counterfactuals that rely heavily on important words (change at least 50\% of them) shows that only Llama-2 benefits from this guidance signal, whereas flip rate decreases for the other two LLMs. This decrease could be the result of two potential reasons: (1) words that are important for the original prediction (label $y$) may not be relevant for counterfactuals, as noted in prior work~\cite{wu2021polyjuice}, and (2) LLMs may fail to provide proper word substitutions. 

\cref{tab:examples} presents examples where LLMs succeed or fail to flip the label. 
In failure cases, despite changing all important words the label remained unchanged, because the words were deleted or replaced by synonyms.
In successful cases, LLMs not only changed important words but also modified other words to improve coherence, resulting in better text quality.

\subsection{More Examples in Few-shot Prompting Do Not Necessarily Improve Performance}
\label{sec:results:few_shot}
\cref{tab:shot_results} shows that increasing the number of shots improves the flip rate on SNLI (except for GPT-4o-mini), as well as distance and perplexity on IMDB, but has a minimal effect on the flip rate for IMDB. The impact of the number of shots varies by dataset and LLM type; for example, Llama-2 shows a strong decrease in distance when the number of shots increases on SNLI. On IMDB and across all other models on both datasets, differences are minimal or non-existent. Additionally, increasing the number of shots can degrade LLM performance, especially for models with smaller context lengths like Llama-2. For instance, the flip rate on IMDB drops significantly with 5 shots, and Llama-2 cannot generate meaningful CFs with 10 shots on IMDB due to exceeding its 4096-token context limit. Therefore, we adopt a 1-shot setting for the representative results, even though n-shot settings may outperform it in certain cases.

\begin{table}[ht!]
\caption{Influence of the number of shots (\# shots) for few-shot prompting. Results for counterfactual metrics (FR: flip rate, Dis:  token-level Levensthein Distance, PP: perplexity) on IMDB and SNLI across different LLMs. \textbf{Bold} denotes the best value across all models, and the second-best value is \underline{underlined}. 10-shot prompting is not applicable (n.a.) for Llama-2 since the prompt is larger than the size of the model's context.}
\label{tab:shot_results}
\centering
\addtolength{\tabcolsep}{4pt} 
\begin{tabular}{ccp{1.3cm}cccccc}
\toprule
\multirow{2}{*}{\textbf{\# shots}}
&&& \multicolumn{3}{c}{\textbf{IMDB}} & \multicolumn{3}{c}{\textbf{SNLI}} \\
\cmidrule(lr){4-6} \cmidrule(lr){7-9}
&&& \textbf{FR} $\uparrow$  & \textbf{Dis} $\downarrow$ & \textbf{PP} $\downarrow$ 
& \textbf{FR} $\uparrow$  & \textbf{Dis} $\downarrow$ & \textbf{PP} $\downarrow$ \\
\midrule
\multirow{9}{*}{1} 
&\multirow{3}{*}{Llama-2}
&CGG& 0.86  & 90.6 & 60 &  0.40 & 6.4 & \underline{54}  \\
&&CGV& 0.94  & 65.5 & \textbf{45} & 0.70 & 4.0 & 56  \\
&&CGGV& 0.97  & 64.0 & 68  & 0.65  & 5.0 & 56  \\
\cmidrule(){2-9}
&\multirow{3}{*}{Llama-3.1}
&CGG& 0.94 & 45.4 & 56 & 0.40 & 4.4  & 58 \\
&&CGV& \underline{0.99} & 34.6 & \underline{48} & 0.84 & 3.7 & 55 \\
&&CGGV& \textbf{1.0} & 32.7& 55 & 0.73 & 3.3  & 59  \\
\cmidrule(){2-9}
&\multirow{3}{*}{GPT-4o-m}
 & \cgg & 0.97 & 43.9 & 54 & 0.44 & 3.8 & 62\\
 && \cgv & \underline{0.99} & 34.5 & 47 & 0.75 & 3.1 & 61\\
 && CGGV & \underline{0.99} & 36.6 & 54 &0.63 & 3.1 & 64\\

\midrule
\multirow{9}{*}{5}
&\multirow{3}{*}{Llama-2}
&CGG& 0.77 & 35.2 & 89 &  0.40 & 5.5 & 56 \\
&&CGV& 0.81 & \textbf{25.0} & 52 & 0.72 & 4.0 & \textbf{53} \\
&&CGGV& 0.86 & 31.2 & 49 & 0.67 & 4.4 & \textbf{53} \\
\cmidrule(){2-9}
&\multirow{3}{*}{Llama-3.1}
&CGG& 0.94 & 38.0 & 49 & 0.38 & 3.8 & 58 \\
&&CGV& \underline{0.99}  & 30.0 & 49  & \underline{0.86} & 3.4 & 58\\
&&CGGV& \underline{0.99}  & 29.3 & 51  & 0.73 & \textbf{2.8} & 61\\
\cmidrule(){2-9}
& \multirow{3}{*}{GPT-4o-m}
 & \cgg & 0.95 & 37.9 & 51 & 0.41 & 3.4 & 64 \\
 && \cgv & 0.98 & 32.7 & 47 & 0.69 & 3.5 & 59 \\
 && CGGV & \underline{0.99} & 31.1 & 50 & 0.53 & 3.1 & 65\\
\midrule
\multirow{6}{*}{10}
&\multirow{3}{*}{Llama-2}
&CGG& n.a. & n.a. & n.a. & 0.41 & 5.8 & 55 \\
&&CGV& n.a. & n.a. & n.a. &   0.75 & 4.0 & 55\\
&&CGGV& n.a. & n.a. & n.a. & 0.66 & 4.5 & \textbf{53}\\
\cmidrule(){2-9}
&\multirow{3}{*}{Llama-3.1}
&CGG&  0.95 & 37.0 & 49 & 0.40 & 3.8 & 60 \\
&&CGV& 0.98 & 29.1 & \underline{48} & \textbf{0.87} & 3.4 & 58  \\
&&CGGV& \textbf{1.0} & \underline{29.0} & 50 & 0.72 & \textbf{2.8} & 61 \\
\cmidrule(){2-9}
& \multirow{3}{*}{GPT-4o-m}
 & \cgg & 0.95 & 36.6& 50&0.42 & 3.3 & 64\\
 && \cgv & 0.98 &32.1&48&0.70 & 3.5 &59\\
 && CGGV & \underline{0.99} &30.1&50&0.53 & \underline{3.0} &65\\
\bottomrule
\end{tabular}
\end{table}

\begin{table}[ht]
    \caption{Data augmentation performance. The original (baseline) training set is augmented with additional data (+) by multiple counterfactual generation methods. Orig is the original test set (in-distribution), Expert and Crowd are human-generated counterfactual test sets (out-of-distribution).
    \textit{Italic} denotes the best value within the same LLM, \textbf{bold} denotes the best value overall, and the second-best value is \underline{underlined}. We report the average accuracy over three runs (standard deviation is less than 0.01).}
    \label{augmentation}
    \centering 
\addtolength{\tabcolsep}{2pt} 
\begin{tabular}{lcl@{\hspace{1cm}}ccccc}
\toprule
%&&&\multicolumn{5}{c}{\textbf{Test}}\\
%\cmidrule{4-8}
\multicolumn{3}{l}{\textbf{Augmentation Setup}}
& \multicolumn{3}{c}{\textbf{IMDB}} & \multicolumn{2}{c}{\textbf{SNLI}} \\
&& & Orig & Expert & Crowd  & Orig &  Crowd  \\
\midrule
\multicolumn{3}{l}{Original (Baseline)}  & 0.91  &   0.88  & 0.93 & 0.74 & 0.50\\
&            &+ Human & 0.90  &   \textbf{0.93}  &  \textbf{0.96} &0.82 &\textbf{0.71}\\ \midrule
  &          &+ FLARE & 0.90  &  0.88   & 0.94 & 0.75 & 0.55\\
   &         &+ MICE & 0.89  &  0.88   & \underline{0.95} & 0.78 & 0.61\\\midrule
%            \multirow{4}{*}{\rotatebox{90}{\makecell{Llama-2}}}
\multirow{4}{*}{Llama-2} &
            &+ vanilla&  0.91  &  0.88   & 0.93 & 0.82 & 0.53 \\
&            &+ CGG&  \textbf{\textit{0.93}} &  0.87   & 0.92 & 0.81 & 0.63 \\
&            &+ CGV&  0.91 & \underline{\textit{0.90}}  & \textit{0.94} & 0.83 & 0.63\\
 &           &+ CGGV &  0.91 &   \underline{\textit{0.90}} &  \textit{0.94} & \underline{\textit{0.84}} & \underline{\textit{0.65}}\\\midrule
%            \multirow{4}{*}{\rotatebox{90}{\makecell{Llama-3.1}}}
\multirow{4}{*}{Llama-3.1} &
            &+ vanilla&  0.91  &  0.86   & 0.92  & \textit{\textbf{0.86}} & 0.55\\
 &           &+ CGG& 0.91  &   0.88  &  0.94 & 0.83 & 0.65\\
 &           &+ CGV& \underline{\textit{0.92}}  & \underline{\textit{0.90}}  & \underline{\textit{0.95}} & 0.75 & 0.54 \\
 &           &+  CGGV & 0.90  &   \underline{\textit{0.90}}  &  0.94 & 0.83 & \underline{\textit{0.65}}\\\midrule
%            \multirow{4}{*}{\rotatebox{90}{\makecell{GPT-4o-m}}}
 \multirow{4}{*}{GPT-4o-m} &           
              &+ vanilla&  \textit{0.91}  &  0.88   & \textit{0.94} & 0.80 & 0.57\\
 &           &+ CGG& \textit{0.91 } &  0.86   & 0.92 & \textit{0.82} & \textit{0.62} \\
  &          &+ CGV&  \textit{0.91} & 0.87  & 0.93 & 0.74 & 0.49\\
  &          &+ CGGV & 0.90  &  \textit{0.89}   & 0.93 & 0.81 & \textit{0.62} \\

%Mixtral 56B\footnote{non-instruct}&  44.42 ± 3.55 &     45.92 ± 4.62 & 55.15 ± 2.57 \\
\bottomrule 
\end{tabular}
\end{table}

\subsection{LLM-Generated CFs Are Helpful for Data Augmentation}
\label{sec:results:augment}
We evaluated whether the generated CFs improve model robustness by following the experimental setup from \cite{Kaushik-etal-2020-learning}. Specifically, we augmented the training data from \cite{Kaushik-etal-2020-learning} with our generated CFs and trained a BERT classifier on the augmented data. We then evaluated the classifier on three test sets: the original test set, a CF test set created through crowdsourcing~\cite{Kaushik-etal-2020-learning}, and a CF test set created by the dataset authors~\cite{gardner-etal-2020-evaluating}. The latter two are considered out-of-distribution data~\cite{gardner-etal-2020-evaluating}, where models typically struggle.
As \cref{augmentation} shows, the CFs generated by our methods improve classifier accuracy on all datasets, except for GPT-4o-mini on the IMDB dataset. On the original test set, our CF-augmented data outperformed human-augmented data. However, on the more challenging human-generated test sets, LLM-generated CFs perform worse than human-augmented data but still outperform models trained without data augmentation and also outperform other counterfactual generation methods such as FLARE and MICE. These results demonstrate that our methods not only enhance CF generation but also improve downstream task performance.

\subsection{LLMs Are Not Faithful to the Classifier, but Rely on Parametric Knowledge}
\label{sec:results:parametric}
When a classifier performs well, its decisions align with real-world logic, making it challenging to determine whether LLMs explain the classifier’s reasoning or rely on real-world knowledge that coincidentally aligns with the classifier’s decisions. Conversely, poorly performing classifiers often deviate from real-world logic. If LLMs can generate effective CFs for such classifiers, it indicates faithfulness to the classifier’s reasoning rather than external knowledge.  
To evaluate this faithfulness, we tested LLMs using a low-accuracy classifier. Specifically, we trained a logistic regression classifier on the IMDB dataset with reversed labels (i.e., positive swapped with negative) while keeping the test set unchanged, resulting in 17\% accuracy. This setup forces LLMs to explain a classifier that operates contrary to their parametric knowledge~\cite{yu-etal-2024-revealing}. If LLMs could successfully explain this classifier, it would demonstrate faithfulness to any classifier, regardless of performance.
%Since the classifier's positive labels are largely negative (and vice versa) based on common sense, this conflicts with the LLMs' parametric knowledge.
%\paragraph{Can LLMs generate counterfactual explanations for any classifiers?}
%To evaluate whether LLMs can truly generate counterfactual explanations for any classifier, we experiment with a low-accuracy model. We train a logistic regression classifier on the IMDB dataset with reversed labels—positive swapped with negative—while leaving the test set unchanged. This results in a classifier with 17\% accuracy on the test set, predicting reversed labels. 

\begin{table}[t]
\caption{Low-accuracy classifier. Results for counterfactual metrics (FR: flip rate, Dis:  token-level Levenshtein Distance, PP: perplexity) on IMDB. $\Delta$ indicates the difference compared to the same setting with the high-accuracy classifier.}
\label{tab:bad_results}
\centering
\addtolength{\tabcolsep}{3pt} 
% \footnotesize
\begin{tabular}{cl@{\hspace{0.8cm}}cllc}
\toprule
&& \textbf{FR} $\uparrow$($\Delta$)  & \textbf{Dis} $\downarrow$($\Delta$)  & \textbf{PP} $\downarrow$($\Delta$)  & \textbf{MR} \\
\midrule
% \multirow{3}{*}{\centering\rotatebox{90}{\makecell{\scriptsize Llama-2}}} 
\multirow{3}{*}{Llama-2} 
&\cgg& 0.16 (-0.81)  & 69.3 (-0.24) & 81 (+0.35) & 0.33  \\
&\cgv& 0.44 (-0.53)  & 44.9 (-0.31) & 61 (+0.36)  & 0.21  \\
&CGGV& 0.38 (-0.60) & 54.8 (-0.14) & 59 (-0.13)  & 0.29  \\
\midrule
%\multirow{3}{*}{\centering\rotatebox{90}{\makecell{\scriptsize Llama-3.1}}}
\multirow{3}{*}{Llama-3.1}
& \cgg & 0.28 (-0.70) & 48.1 (+0.06) & 67 (+0.19)  & 0.47  \\
& \cgv & 0.76 (-0.23) & \textbf{31.1} (-0.1) & 49 (+0.02) & 0.19   \\
& CGGV& 0.46 (-0.54) & 39.0 (+0.19) & 72 (+0.31) & 0.42  \\
\midrule
%\multirow{3}{*}{\centering\rotatebox{90}{\makecell{\scriptsize GPT-4o-m\\}}}
\multirow{3}{*}{GPT-4o-m}
&\cgg & 0.38 (-0.61)  & 45.7 (+0.04) & 57 (+0.06)  &0.53 \\
&\cgv& \textbf{0.79} (-0.20) & \textbf{31.1} (-0.1) & \textbf{47} (0) & 0.19  \\
&CGGV& 0.54 (-0.45)  & 40.2 (+0.1) & 57 (+0.06)  & 0.49\\
\bottomrule\\
\end{tabular}
% \addtolength{\tabcolsep}{2.5pt}
\end{table}

\cref{tab:bad_results} shows that LLMs struggle to explain the low-accuracy classifier, with a significantly lower FR compared to the high-accuracy classifier. This suggests that LLMs heavily rely on their parametric knowledge, despite being prompted to adapt to reversed labels, aligning with prior findings in other tasks~\cite{chen-etal-2022-rich,longpre-etal-2021-entity}. GPT-4o-mini performs best, reflecting better instruction understanding, though it still leans on parametric knowledge. \cgv{} is the most robust method, maintaining moderate flip rates on Llama-3.1 and GPT-4o-mini, while Llama-2 and \cgg{} show sharp declines, indicating an over-reliance on parametric knowledge. Distance and perplexity also worsen, further illustrating the difficulty of explaining low-accuracy classifiers.  
These results emphasize the importance of evaluating LLM-based counterfactual generation on both high- and low-accuracy classifiers. While low-accuracy classifiers are unlikely to be deployed in practice, they provide a proxy for wrong classifier decisions and more importantly, allow us to evaluate the faithfulness of counterfactual explanations, i.e., whether counterfactuals align with the classifier's true reasoning.
%Further research is needed to improve LLMs' focus on context over parametric knowledge.

% \begin{table}[t]
%     \centering 

% \begin{tabular}{lccccc}

% \toprule
% &\multicolumn{5}{c}{\textbf{Test Data}}\\
% \cmidrule{2-6}
% & \multicolumn{3}{c}{\textbf{IMDB}} & \multicolumn{2}{c}{\textbf{SNLI}} \\
% & Orig & Expert & Crowd & Orig & Crowd   \\
% \midrule
%            Original & 0.91  &   0.88  & 0.93 & 0.74 & 0.50\\
%             + Human & 0.90  &   \textbf{0.93}  &  \textbf{0.96} &0.82 &\textbf{0.71}\\ \midrule
%             + Llama-2 & \textbf{0.93}  & \underline{0.90}    & 0.94 &\textbf{0.84} & \underline{0.65} \\
%             + Llama-3 & \underline{0.92}  &  \underline{0.90}   & \underline{0.95} & \underline{0.83} &\underline{0.65}\\
%             + GPT-4o-m & 0.91  &  0.89   & 0.93 & 0.82 & 0.62 \\

% %Mixtral 56B\footnote{non-instruct}&  44.42 ± 3.55 &     45.92 ± 4.62 & 55.15 ± 2.57 \\
% \bottomrule

% \end{tabular}

%     \caption{Data augmentation results with \textit{Accuracy} metric. For each LLM, we report the best-performing approach as the representative result. Full details and the complete table are provided in \cref{appx:augmentation}.}
%     \label{tab:augmentation}
% \end{table}

\section{Conclusion}
% With the proposed approaches, \cgg{} and \cgv{}, we demonstrate that integrating a classifier into counterfactual generation improves counterfactual metrics while preserving the strengths of LLMs in text quality. However, identifying important words for counterfactual explanations, which play a critical role in \cgg{}, remains a challenge as the current XAI methods only focus on important words for the prediction. Additionally, the reliance of LLMs on parametric knowledge requires further investigation to improve their faithfulness to the classifiers.
In this paper, we presented two simple yet effective approaches that use information from the classifier to guide counterfactual generation by LLMs.
Our approaches  improve the performance of multiple LLMs on counterfactual metrics by such margins that some even surpass state-of-the-art counterfactual generation methods, highlighting the value of leveraging classifier information.
With our proposed approaches, we emphasize the important role of classifier information in improving counterfactual generation by LLMs and encourage its use for explanations or data augmentation in the future. Our experiments show that using the classifier's labels for validation is the most effective guidance to improve counterfactual metrics, while using important words as a guidance signal requires further research. 
We further show that augmenting the training data of a classifier with counterfactuals generated by our approach can improve model accuracy and robustness.

Our analysis uncovers a critical issue in counterfactual explanations by LLMs: LLMs are not fully faithful to the classifier, often relying more on parametric knowledge than the classifier's actual reasoning. Developers of future LLM-based counterfactual generation methods should pay attention to this issue. We strongly recommend evaluating counterfactuals with both high- and low-accuracy classifiers to ensure the faithfulness of counterfactual explanations.
%Finally, we show that the CFs generated by our method can improve model accuracy and model robustness by augmenting the training data set with the generated counterfactuals.

% Although \cgg{} depends on the internal knowledge and type of LLM, it effectively guides LLMs to leverage classifier information in generating counterfactuals. In contrast, \cgv{} is a simple yet effective approach in most cases.
% \clearpage
% \section{Limitations}
% Data leakage could be an issue, as LLMs are known to be exposed to datasets from humans~\cite{balloccu-etal-2024-leak}, which might include human-generated counterfactuals. However, \cite{nguyen-etal-2024-ceval-benchmark} showed  that counterfactuals generated by LLMs differ from those in human-generated counterfactual datasets. 
% We did not excessively tune the prompts to generate counterfactual methods. Prompting techniques, such as Chain-of-Thought~\cite{wei2022chain} or self-consistency~\cite{wang2023selfconsistency} could maybe further improve performance, but since LLMs are sensitive to prompts~\cite{lu-etal-2022-fantastically}, excessive tuning might overfit the datasets under consideration.
%Another limitation is the optimization of the prompts; some prompting techniques, such as Chain-of-Thought~\cite{wei2022chain} or self-consistency~\cite{wang2023selfconsistency}, may improve performance further. 

% \bibliography{anthology, custom}
\bibliographystyle{splncs04}
\bibliography{anthology,custom}

\appendix

\section{Example Prompts}
\label{appx:prompts}
\cref{fig:prompt:gen:sent} and \cref{fig:prompt:gen:snli} present example prompts for sentiment analysis and NLI tasks used in the CGG approach.

\begin{figure*}[h]
\setlength\fboxsep{7pt}
\fbox{%
\parbox{0.95\linewidth}{%
\begin{courier}
Given a movie review with its original sentiment classified as either positive or negative by a classifier, your task is to modify the text with minimal edits to flip the sentiment prediction of the classifier. Please enclose the generated text within <new> and </new> tags\\
You will be provided with a list of important words identified by the classifier as influential in its prediction. Use these words as a guide for your changes, ensuring that the\\
 text remains fluent. Avoid making any unnecessary changes.\\
Example:\\
Positive: Long, boring, blasphemous. Never have I been so glad to see ending credits roll.\\
Target sentiment: Negative\\
You must change these following important words: ['boring,', 'blasphemous.', 'glad'] to achieve the target sentiment.\\
Generated Text:\\
Negative: <new>Long, fascinating, soulful. Never have I been so sad to see ending credits roll.</new>\\
Request: Similarly, given a movie review with its original sentiment classified as either positive or negative by a classifier, your task is to modify the text with minimal edits to flip the sentiment prediction of the classifier.\\
Positive: A spoiler.  What three words can guarantee you a terrible film? Cheap Canadian Production. THE BRAIN fits those words perfectly. Terrible script, idiotic acting and hilarious special effects make this a must for every BAD movie fan. The horror is hilarious. The post production team looks like it gave up. What makes THE BRAIN admirable is in the second half, it actually tries to be good! Can a bit of ingenuity and consistency save what is already a joke?  It's around Christmas time. A mother and daughter are murdered by one of the funniest looking villains ever. The day later, a rebel teen gets into enough trouble that he is sent for a psychiatric analysis.  If a cop 's head is chopped off and a stranger with blood on him and a bloody axe told you some kids did it, who would you believe? What begins as funny turns dull and tiring toward the end when THE BRAIN tries to be serious. A child cannot be frightened by the scary moments. THE BRAIN is too funny a concept to try and be gritty. The Psychological Research Institute is larger than major manufacturing plants! Our ugly villain and its cohorts get credit for pulling some of the worst acting I have seen. Viewer discretion advised heavily.\\
Target sentiment: Negative\\
You must change these following important words: ['worst', 'villain', 'dull', 'murdered', 'save', 'idiotic', 'concept', 'tries', 'tries', 'guarantee', 'terrible', 'joke', 'stranger', 'post', 'ugly', 'looking', 'looks', 'script', 'child', 'blood', 'scary', 'half', 'teen', 'fits', 'gave', 'told', 'special', 'horror', 'trouble', 'did', 'kids', 'funny', 'funny', 'make', 'daughter', 'cop', 'mother', 'believe', 'major', 'begins', 'gets', 'turns', 'end', 'axe', 'advised', 'movie', 'chopped', 'time', 'plants', 'like'] to achieve the target sentiment.\\
You must enclose your generated text within opening '<new>' and closing '</new>' tags, like the examples above. No further explanations are needed. Your generated text:

\end{courier}
}}
\caption{Prompt for CFs generation - IMDB - CGG}
\label{fig:prompt:gen:sent}
\end{figure*}
\begin{figure*}
\setlength\fboxsep{7pt}
\fbox{%
\parbox{0.95\linewidth}{%
\begin{courier}
Context: For NLI task, given two sentences (premise and hypothesis) and their original relationship, determine whether they entail, contradict, or are neutral to each other. Change the premise with minimal edits to achieve the neutral relationship from the original one. Do not make any unnecessary changes.\\
You will be provided with an ordered list of important words identified by the classifier as influential in its prediction (First word is the most important). Use these words as a guide for your changes.\\
Do not make any unnecessary changes. Make as few edits as possible. You must enclose your generated text within opening '<new>' and closing '</new>' tags, like the examples above. No further explanations are needed.\\
For example:\\
Original relationship: entailment\\
Premise: Seven people are racing bikes on a sandy track.\\
Hypothesis: The people are racing.\\
Target relationship: neutral\\
You must change these following words: ['racing'] in the premise to achieve the target relation . \\
(Edited premise): <new>Seven people are riding bikes on a sandy track.</new>

Request: Similarly, given two sentences (premise and hypothesis) below and their original relationship. Change the premise with minimal edits to achieve the target neutral relationship.\\
You will be provided with an ordered list of important words identified by the classifier as influential in its prediction (First word is the most important). Use these words as a guide for your changes.\\
Original relationship: entailment\\
Premise: A man with a beard is talking on the cellphone and standing next to someone who is lying down on the street.\\
Hypothesis: A man is prone on the street while another man stands next to him.
Target relationship: neutral\\
You must change these following words: ['cellphone', 'lying', 'beard', 'talking'] in the premise to achieve the target relation . \\
Do not make any unnecessary changes. Make as few edits as possible. You must enclose your premise within opening '<new>' and closing '</new>' tags, like the examples above. No further explanations are needed. Only return the premise.
(Edited premise):

\end{courier}
}}
\caption{Prompt for CFs generation - SNLI - CGG}
\label{fig:prompt:gen:snli}
\end{figure*}

\section{SHAP as Feature Importance Method}
\label{appx:shap}
The results for \cgg{} and CGGV using SHAP~\cite{lundberg2017shap} in \cref{appx:tab:shap_results} indicate that using SHAP as the XAI method to identify important words performs similarly as saliency maps, reinforcing a limitation of XAI methods: words important for the current prediction are not necessarily important for counterfactual generation.
\begin{table}[th!]
\caption{SHAP as feature extractor. Results for counterfactual metrics (FR: flip rate, Dis:  token-level Levensthein Distance, PP: perplexity) on IMDB and SNLI across different LLMs.  \textbf{Bold} denotes the best value across all settings.}
\label{appx:tab:shap_results}
\centering
\addtolength{\tabcolsep}{3pt} 
\begin{tabular}{cp{1.6cm}@{\hspace{0.5cm}}cccccc}
\toprule
&& \multicolumn{3}{c}{\textbf{IMDB}} & \multicolumn{3}{c}{\textbf{SNLI}} \\
\cmidrule(lr){3-5} \cmidrule(lr){6-8}
&& \textbf{FR} $\uparrow$  & \textbf{Dis} $\downarrow$ & \textbf{PP} $\downarrow$ 
& \textbf{FR} $\uparrow$  & \textbf{Dis} $\downarrow$ & \textbf{PP} $\downarrow$ \\
\midrule
\multirow{3}{*}{CGG}
&Llama-2&  0.85 & 55.8 & 65 & 0.29 & 6.2 & \textbf{56} \\
&Llama-3.1& 0.94 & 39.7 & 51 & 0.28 & 4.0 & 60   \\
&GPT-4o-m& 0.98 & 38.0 & 50 & 0.32 & 3.3 & 65 \\
\midrule
\multirow{3}{*}{CGGV}
&Llama-2& 0.95 & 33.6 & \textbf{44 }& 0.53 & 4.7 & 61   \\
&Llama-3.1& 0.99 & \textbf{30.2 }& 51 &\textbf{ 0.62} & 3.0 & 63    \\
&GPT-4o-m& \textbf{1.00} & 31.2 & 52 &0.53 & \textbf{2.9} & 66  \\
\bottomrule
\end{tabular}
\end{table}

\end{document}